\documentclass[runningheads]{llncs}

\usepackage{eccv}

\usepackage{eccvabbrv}

\usepackage{graphicx}
\usepackage{booktabs}
\usepackage{placeins}
\usepackage{tcolorbox}
\usepackage{colortbl}
\usepackage{lipsum}
\usepackage{multirow}
\usepackage{pifont}
\usepackage{wrapfig}
\usepackage{makecell}
 
\usepackage[accsupp]{axessibility}  

\usepackage[pagebackref,breaklinks,colorlinks,citecolor=eccvblue]{hyperref}

\usepackage{orcidlink}

\begin{document}

\title{7ABAW-Compound Expression Recognition via Curriculum Learning} 

\titlerunning{Compound Expression Recognition via Curriculum Learning}
\author{Chen Liu\inst{1,2}\textsuperscript{*} \orcidlink{0000-0003-3159-0034} \and
Feng Qiu\inst{1}\textsuperscript{*} \orcidlink{0000-0002-4333-5957} \and
Wei Zhang\inst{1} \orcidlink{0000-0001-5907-7342}\and
Lincheng Li\inst{1}\textsuperscript{$\dagger$} \orcidlink{0000-0002-6047-0472} \and
Dadong Wang\inst{3}\orcidlink{0000-0003-0409-2259} \and Xin Yu\inst{2}\orcidlink{0000-0002-0269-5649}
}

\authorrunning{C.~Liu et al.}

\institute{NetEase Fuxi AI Lab, Hangzhou, China \\
\email{\{qiufeng, zhangwei05, lilincheng\}@corp.netease.com} \and
The University of Queensland, Queensland, Australia\\ 
\email{chen.liu7@uqconnect.edu.au, xin.yu@uq.edu.au} 
\and 
CSIRO, Data61, Sydney, Australia\\
\email{Dadong.Wang@data61.csiro.au}
\footnote[1]{* Equal Contribution, $\dagger$ Corresponding authors. This work is done at Netease.}
}

 
\maketitle

\begin{abstract}
With the advent of deep learning, expression recognition has made significant advancements. 
However, due to the limited availability of annotated compound expression datasets and the subtle variations of compound expressions, Compound Emotion Recognition (CE) still holds considerable potential for exploration. 
To advance this task, the 7th Affective Behavior Analysis in-the-wild (ABAW) competition introduces the Compound Expression Challenge based on C-EXPR-DB, a limited dataset without labels.
In this paper, we present a curriculum learning-based framework that initially trains the model on single-expression tasks and subsequently incorporates multi-expression data. 
This design ensures that our model first masters the fundamental features of basic expressions before being exposed to the complexities of compound emotions. 
Specifically, our designs can be summarized as follows:
1) \textbf{Single-Expression Pre-training:} The model is first trained on datasets containing single expressions to learn the foundational facial features associated with basic emotions.
2) \textbf{Dynamic Compound Expression Generation:} Given the scarcity of annotated compound expression datasets, we employ CutMix and Mixup techniques on the original single-expression images to create hybrid images exhibiting characteristics of multiple basic emotions. 
3) \textbf{Incremental Multi-Expression Integration:} After performing well on single-expression tasks, the model is progressively exposed to multi-expression data, allowing the model to adapt to the complexity and variability of compound expressions.
The official results indicate that our method achieves the \textbf{best} performance in this competition track with an F-score of 0.6063. Our code is released at \href{https://github.com/YenanLiu/ABAW7th}{this repository}.
\end{abstract}    
\section{Introduction}
\label{sec:intro}
The Compound Expression (CE) Recognition task focuses on identifying complex emotional states conveyed by facial expressions that are combinations of basic emotions. Unlike recognizing single, basic emotions such as happiness, sadness, or anger, compound expressions involve more nuanced and mixed emotions, such as happily surprised, sadly angry, or fearfully disgusted. These compound expressions provide a richer and more accurate representation of human affective states\cite{kollias20246th, kollias2020analysing, yin2023multi, kollias2021analysing, zhang2023multi, nguyen2023transformer, ritzhaupt2021meta, kollias2023abaw2, kollias2023abaw, kollias2021distribution, kollias2019expression, kollias2019deep, kollias2019face, zafeiriou2017aff,zhang2024effective, zhang2024affectivebehaviouranalysisintegrating, liu2024affectivebehaviouranalysisprogressive}.

Single-expression recognition has seen significant advancements with the advent of deep learning, which enables more accurate and efficient recognition of basic emotional states. 
However, understanding compound expressions still faces great challenges.
We categorize several challenges inherently of this task as follows: a) \textbf{Limited datasets.}  The annotated compound expression data is very rare, posing a challenge for training robust models.
b) \textbf{Subtle Differences of Compound Expressions.} The compound expression usually contains subtle variations in facial features. Hence, they are more difficult to distinguish than basic emotions.
To facilitate the development of this field, the 7th Affective Behavior Analysis competition (ABAW7) \cite{kollias20247th} set the Compound Expression (CE) \cite{dong2024bi, he2022compound, she2021dive, wang2020suppressing} Challenge based on the C-EXPR-DB \cite{kollias2023multi} dataset.
Participations are required to achieve compound expression recognition in videos with limited amounts and unknown labels \cite{dong2024bi, he2022compound, she2021dive, wang2020suppressing}.

To solve the above challenges, we introduce a curriculum learning framework, which enhances the generalization ability of our model in compound expression recognition tasks by gradually transitioning from basic expression to multi-expression learning.
More specifically, we first train our model on the datasets just containing single expressions.
In this fashion, our model learns the foundational facial features associated with basic emotions.
After achieving satisfactory performance on the single-expression recognition task, we expose the model to data with multiple expressions progressively. 
This stepwise introduction allows the model to adapt to the increased complexity and variability inherent in compound expressions.

To address the scarcity of annotated compound expression datasets, we apply CutMix and Mixup to the original single-expression images, creating hybrid images that exhibit characteristics of multiple basic emotions. The CutMix technique involves cutting and pasting patches from different images, while Mixup generates linear interpolations between pairs of images. These augmented images are then used to train the model, providing it with a rich and varied training set that includes a wide range of compound expressions.
We conduct comprehensive evaluations on the officially provided dataset and our self-curated validation dataset. 
The results demonstrate the effectiveness of our method.

\section{Related Work}
\label{sec:related}

Compared to compound expression recognition, basic expression recognition has achieved many advancements in recent years \cite{guo2018dominant, houssein2022human, canal2022survey, yue2019survey}.
Single-expression recognition has been a foundational area of research in the field of affective computing and computer vision \cite{zhang2022learn, liu2023audio, liu2023bavs, qi2023diverse, qi2023emotiongesture}. Early methods rely heavily on handcrafted features such as Local Binary Patterns (LBP) \cite{pietikainen2010local}, Histogram of Oriented Gradients (HOG) \cite{dalal2005histograms}, and Gabor filters \cite{movellan2002tutorial} to capture the distinct characteristics of facial expressions. These traditional techniques, though effective to some extent, faced limitations in handling variations in lighting, pose, and occlusion.

With the advent of deep learning, particularly convolutional neural networks (CNNs), the performance of single expression recognition systems has seen significant improvements  \cite{yu2017hallucinating, yu2017face, yu2018face, yu2019can, yu2018super}. CNNs automatically learn hierarchical feature representations from raw pixel data, leading to more robust and accurate recognition.  
Hasani and Mahoor \cite{hasani2017facial} employ deep residual networks for CER, while Kosti \emph{et al.} \cite{kosti2017emotion} use LSTM networks to capture temporal dynamics in video sequences. Multi-task learning (MTL) frameworks, as explored by Zhang \emph{et al.} \cite{zhang2022multi}, have also enhanced CER by leveraging shared feature extractors with task-specific heads. 

Further advancements were made by incorporating transfer learning and fine-tuning pre-trained models on large-scale face datasets. This approach leveraged the generalization capabilities of models trained on extensive datasets, such as VGG-Face \cite{qawaqneh2017deep} and ResNet \cite{li2021facial}, to enhance the performance of expression recognition tasks. 
Additionally, researchers have explored the integration of attention mechanisms and ensemble learning to refine the focus on critical facial regions and combine the strengths of multiple models \cite{savchenko2021facial, savchenko2022classifying, zhang2022learn}, respectively.
However, given the complexity of human emotions in real-world situations, detecting a single expression is inadequate.

Compound Expression Recognition (CER) extends traditional emotion recognition by identifying complex, blended expressions. 
Early efforts by \cite{dong2024bi, he2022compound} systematically categorized compound emotions, laying the groundwork for further studies \cite{yu2017hallucinating, yu2017face, yu2018face, yu2019can, yu2018super}. 
Dimitrios \cite{kollias2023multi} curate the Multi-Label Compound Expression dataset, C-EXPR, and introduced C-EXPR-NET, which simultaneously tackles Compound Expression Recognition (CER) and Action Unit (AU) detection tasks. 
Despite progress, challenges remain due to the subtlety and variability of compound expressions, highlighting the requirement for more sophisticated models and diverse datasets.


\section{Method}
\label{sec:method}
\begin{figure*}[t]
  \centering
\includegraphics[width=1\linewidth]{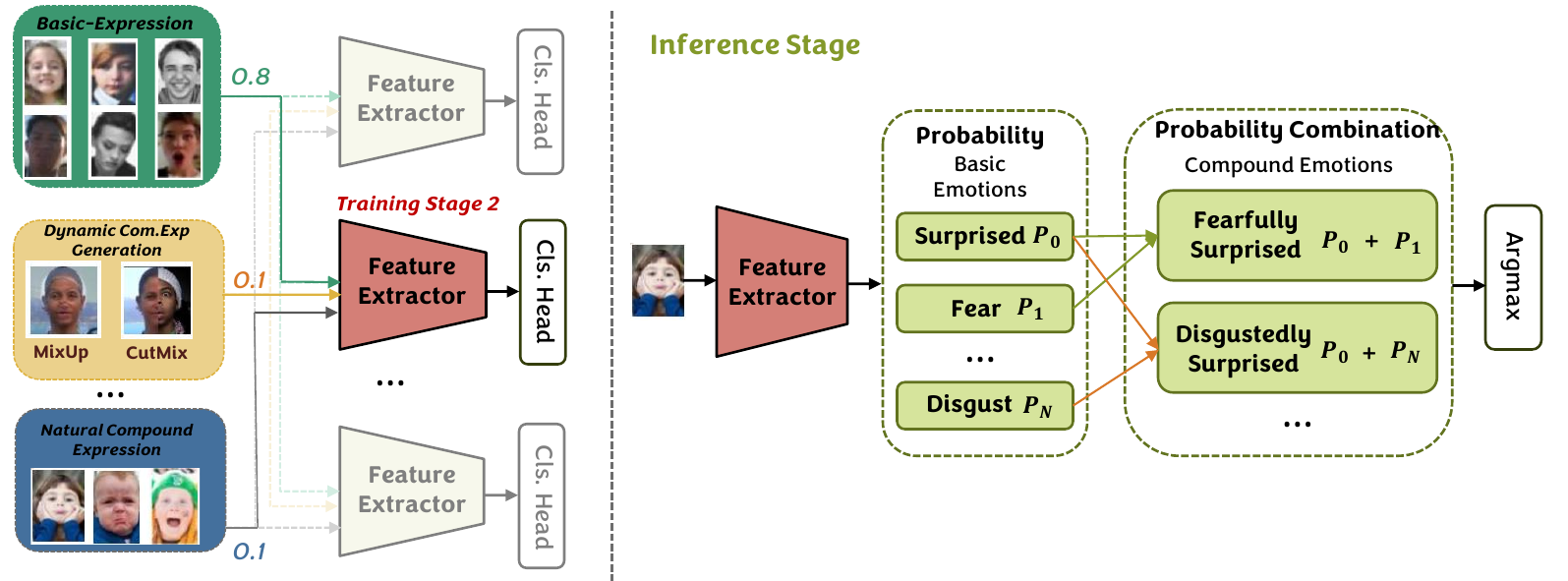}
\caption{\textbf{Illustration of our proposed frameworks for the compound recognition competition.} We adopt a curriculum learning approach, transitioning from basic expression prediction to compound expression learning. We take the second stage training process as an example to illustrate the transition process. Different from the first training stage only utilizes the basic expression data to train the model, in this second stage, we randomly select the compound expression data from the natural compound expression datasets 
(\emph{i.e.} RAD-BF and Fuxi-EXPR ) and the generated compound expression data.  Specifically, there are 80\% basic expression images and 20\% compound expression images involved in the second training stage. Here, cls. Head refers to the classification head, which is composed of linear layers.}
  \label{fig:pipeline}
\end{figure*}

In this work, we employ Masked Autoencoder (MAE)~\cite{he2022masked} as our feature extractor.
In this section, we present our approach for the two competition tracks in three parts.
First, we will describe the construction process of the feature extractor utilized in this challenge.
Then we detail our framework flow (as depicted in Fig. \ref{fig:pipeline}) for the Compound Expression Recognition (CER) challenge.
 
\subsection{Method Overview}
In our approach, we begin by training the model on single expression recognition tasks. 
This involves learning to identify basic emotions such as Neutral, Anger, Disgust,  Fear, Happiness, Sadness, and Surprise. 
Once the model achieves satisfactory performance on the single expression recognition task, we introduce compound expression recognition. 
This progressive training strategy decreases the difficulty of tasks. 
It allows the model to adapt to the more complex task of identifying blended or compound expressions.
Moreover, in the second learning stage, we utilize Cutmix \cite{yun2019cutmix} and Mixup \cite{zhang2017mixup} data augmentation techniques to generate compound expressions.
This increases dataset diversity, enhancing the robustness and generalization of our model.

\subsection{Feature Extractor}

In this competition, we employ Masked AutoEncoder as our feature extractor due to its powerful feature extraction capabilities. 
To obtain high-quality facial features, we first integrate a large-scale facial dataset, including AffectNet \cite{mollahosseini2017affectnet}, CASIA-WebFace \cite{CASIA-Webface}, CelebA \cite{CelebA}, IMDB-WIKI \cite{IMDB-WIKI}, and WebFace260M \cite{zhu2021webface260m}.
The data undergoes further cleaning by the annotation platform to ensure its quality, and ultimately, 4.5 million images are used in the training process.
Based on this dataset, we train the MAE in a self-supervised manner. 
Specifically, we randomly mask 75\% of each facial image and then reconstruct the original face image. 
During the training stage, we utilize L2 Loss to calculate the difference between the reconstructed image and the original image for model optimization. 
To better adapt the MAE model to the affective behavior analysis, we fine-tuned the model on the AffectNet and data from Aff-Wild provided by the competition. 
 
\subsection{Single Expression Recognition}
To attain the high-quality feature extractor, we first finetune the MAE model on RAF-DB and our curated Fuxi-EXPR dataset (including 25K images with seven basic expressions as RAF-DB \cite{li2017reliable}).
Note that, in this stage, we just utilize the data with the single expression label to train the model, as shown in Fig. \ref{fig:pipeline}.
To increase the diversity of the training dataset, we adopt the cutout data augmentation technique as well as the basic data augmentation techniques such as random flipping, color jittering, and random crop in the initial learning stage.
Note that considering the final compound expressions do not include the neutral label, we filter out all data with the neutral label from the entire dataset.

\subsection{Compound Expression Recognition}
As the model's performance stabilizes, we progressively introduce more complex compound expressions to train our model. 
To reduce the risk of overwhelming the model with difficult data too early, we progressively increase the proportion of compound expressions in the training data.
Specifically, we divide the overall training stage into four sub-stages, with the number of training epochs for each stage being 5, 5, 3, and 3. The proportion of compound expressions in each stage is 0, 0.2, 0.4, and 1, respectively.
During the training phase, we treat the compound expression recognition task as a multi-class classification problem. 
To achieve better performance, during the inference phase, we constrain the final results into the officially provided compound expression categories. 
Specifically, we sum the classification probabilities of the individual expression categories involved in the compound expressions and utilize the argmax of these summed probabilities as the final result.

\subsection{Training Objectives}

In the whole training stage, we utilize Binary Cross-Entropy Multi-Label Loss as the optimization objective, mathematically expressed by:
\begin{equation}
\mathcal{L}_{BMC} = -\frac{1}{N} \sum_{i=1}^{N} \sum_{c=1}^{C} \left[ y_{ic} \log(p_{ic}) + (1 - y_{ic}) \log(1 - p_{ic}) \right].
\end{equation}
Here, $N$ is the number of samples, $C$ is the number of categories, $y_{ic}$ is the binary indicator (0 or 1) if class label $c$ is the correct classification for sample $i$, and $p_{ic}$ is the predicted probability of sample 
$i$ being of class $c$.
\section{Experiments}
\label{sec:exper}

\subsection{Dataset}

\subsubsection{Official Provided Data.}
For the CE track, the organizers provide 56 video data from the C-EXPR-DB \cite{kollias2023multi} database. 
These videos are categorized into the following categories: Fearfully Surprised, Happily Surprised, Sadly Surprised, Disgustedly Surprised, Angrily Surprised, Sadly Fearful, and Sadly Angry. 
Note that we do not have the ground truth labels for these videos.

\subsubsection{Extra Data for CE.}
Except for the official provided data, we enhance the dataset with RAF-DB \cite{li2017reliable} and AffectNet \cite{mollahosseini2017affectnet} to increase the data of Anger, Disgust, and Fear expressions.
Fuxi-EXPR contains 71,618 facial images collected from films, television works, and public video platforms. 
Data cleaning and management are handled by the Netease Fuxi Youling Crowdsourcing platform\footnote[2]{https://fuxi.163.com/solution/data}. The dataset includes labels for six basic emotions (Surprised, Fear, Disgust, Happiness, Sadness, Anger) and an additional category for Others.
To obtain stable performance, we retain only the six basic emotions for the two datasets.

\subsubsection{Extra Data for MAE.}
To attain a powerful expression feature extractor, we train MAE on our re-collected facial image dataset, which includes AffectNet \cite{mollahosseini2017affectnet}, CASIA-WebFace \cite{CASIA-Webface}, CelebA \cite{CelebA}, IMDB-WIKI \cite{IMDB-WIKI}, WebFace260M \cite{zhu2021webface260m}, and the Fuxi-EXPR private dataset.

\subsection{Metrics}

For the CE track, the performance $P$ is measured by the average F1 score across all seven categories, expressed by:
\begin{equation}
    P = \frac{\sum_{c=1}^{7} F1_{c}}{7}.
\end{equation}

\subsection{Implementation Details}
All training images are resized to 224 $\times$ 224 pixels. The MAE pertaining is conducted on 8 NVIDIA A30 GPUs with a batch size of 4096.
We conduct the compound expression recognition experiments on 4 NVIDIA 4090 with a batch size of 256.
The training configurations (including the batch size, optimizer, scheduler, learning rate, and so on) can be found in our released code.
Moreover, we divide the training into three stages, with the number of epochs for each stage being 5, 5, 3, and 3. 
The proportion of compound expression images in each stage was 0, 0.2, 0.4, and 1, respectively.

\subsection{Experimental Results}

\begin{table}
\centering
\caption{\textbf{Analysis of Data Augmentation.} All experiments are conducted under the optimal setting of Epoch Distribution and Compound Data Proportion Distribution. The blended weight of the Mixup is set to 0.1. }
\label{tab:aug}
\setlength{\tabcolsep}{2.5 em}
\renewcommand\arraystretch{1.5}
\scriptsize
\begin{tabular}{lllc}
\toprule
Exp & Mixup & CutMix & P      \\ \midrule
1   & \ding{51}                        &        & 0.6764 \\
2   &                           & \ding{51}     & 0.6722 \\
3   & \ding{51}                        & \ding{51}     & 0.6856 \\ \bottomrule
\end{tabular}
\end{table}

\begin{table}
\centering
\caption{\textbf{Analysis of Curriculum Learning.} Epoch Dis. (Epoch Distribution) without numerical values indicates training with only single-label data. Here, Epoch Dis. indicates the epoch number of each training stage. Compound Prop (Compound Data Proportion Distribution) is the proportion of the compound data in each training stage.}
\label{tab:cl}
\setlength{\tabcolsep}{2.3 em}
\renewcommand\arraystretch{1.5}
\scriptsize
\begin{tabular}{c|ll|c}
\toprule
Exp & Epoch Dis.          & Compound Prop.        & P      \\ \midrule
1   & \_                  &  \_                   & 0.6653 \\
2   & {[}5, 15{]}         & {[}0,1{]}             & 0.6759 \\
3   & {[}5, 5, 5{]}       & {[}0,0.5,1{]}         & 0.6652 \\
4   & {[}5, 5, 3, 3{]}    & {[}0,0.2,0.4,1{]}     & \textbf{0.6856} \\
5   & {[}5, 3, 3, 3, 3{]} & {[}0,0.2,0.4,0.6,1{]} & 0.6325 \\ \bottomrule
\end{tabular}

\end{table}

\subsubsection{Analysis of Curriculum Learning.}
We evaluate our CE recognition model with different epoch distributions and compound data proportion distributions, and the results are presented in Table \ref{tab:cl}.
As indicated in Table \ref{tab:cl}, the best performance is achieved under the Epoch distribution of [5, 5, 3, 3] and Compound Propotion distribution of [0, 0.2, 0.4, 1], with the F1 score of 0.6856.
Moreover, we found that the model performance does not always improve with an increasing number of training stages. 
When we divide the training process into five stages and adjust the distribution of compound data to be more even across these stages, the performance decreased by 5.31\% compared to the four-stage training process.

\subsubsection{Analysis of Data Augmentation.}
We evaluate the model performance when incorporating different data augmentation methods in the training stage.
As shown in Table \ref{tab:aug}, the introduction of Mixup and CutMix methods both enhance the model performance. 
Incorporating both data augmentation methods simultaneously achieves the highest F1-score value.

\subsubsection{Evaluation on Official Test Data.}

Table \ref{tab:final} presents the results of all participating teams' approaches on the official test set. 
The results suggested that our approach achieves a significant victory in the compound expression track, outperforming the second place by 0.282.
This also demonstrates that our curriculum learning approach, progressing from simple to complex, facilitates the model to gradually understand and master basic facial features, thereby performing better in handling complex multi-expression tasks.

\begin{table}[]
\centering
\caption{\textbf{The final evaluation results on the official test dataset.} Our team (\textbf{\emph{i.e.} Netease Fuxi AI Lab}) attains the best performance of the C-EXPR-DB dataset.}
\label{tab:final}
\setlength{\tabcolsep}{2.8 em}
\renewcommand\arraystretch{1.5}
\scriptsize
\begin{tabular}{cc}
\hline
Teams               & F1 Score \\ \hline
AIPL-BME-SEU  \cite{li2024temporallabelhierachicalnetwork}      & 0.1644   \\
HFUT-MAC2    \cite{savchenko2024hsemotionteam7thabaw}       & 0.2281   \\
ETS-LIVIA \cite{richet2024textfeaturebasedmodelscompound}          & 0.2591   \\
HSEmotion   \cite{liu2024compoundexpressionrecognitionmulti}        & 0.3243   \\
\textbf{Netease Fuxi AI Lab} (Ours) \cite{liu2024affectivebehaviouranalysisprogressive} & \textbf{0.6063}   \\ \hline
\end{tabular}
\end{table}
\section{Conclusion and Discussion}
\label{sec:Conclu}

\textbf{Conclusion.}
In this paper, we present a curriculum learning-based framework for the 7th ABAW Compound Expression (CE) Recognition task. 
By initially training the model on single-expression images, our model learns the foundational facial features associated with basic emotions. 
Furthermore, we create hybrid images that exhibit characteristics of multiple basic emotions, mitigating the issue caused by the scarcity of annotated compound expression datasets and thus improving the generalization ability of our model. 
Additionally, the model is progressively exposed to multi-expression data, allowing it to adapt to the complexity and variability of compound expressions.
Extensive experiments demonstrate the superiority of our method compared to other solutions.
\vspace{1.0em}

\noindent\textbf{Discussion.}
Although our curriculum learning-based method attains competitive results, there are still many challenges that remain to be addressed.
One of the critical challenges in compound expression recognition is the scarcity of annotated datasets. 
To overcome this, our method includes dynamic compound expression generation techniques such as CutMix and Mixup, which synthesize new training data by combining single-expression images. 
This approach not only expands the training dataset but also introduces variability, which is crucial for training a model capable of handling the diversity of compound expressions.
Moreover, integrating multimodal data, such as audio, physiological signals, and contextual information, could further enhance the accuracy and reliability of compound expression recognition systems. This multimodal approach would provide a more holistic understanding of human emotions, leading to applications that can better interpret and respond to human affective states in real time.

\vspace{1.0em}
\noindent\textbf{Acknowledgements.} This research is funded in part by the National Key R\&D Program of China (No. 2022YFF09022303 to Lincheng Li), ARC-Discovery grant (DP220100800 to XY), and ARC-DECRAgrant (DE230100477 to XY). The first author is funded by the CSC scholarship and Data61-topup scholarship.
 
\bibliographystyle{splncs04}
\bibliography{main}

\begin{thebibliography}{10}
\providecommand{\url}[1]{\texttt{#1}}
\providecommand{\urlprefix}{URL }
\providecommand{\doi}[1]{https://doi.org/#1}

\bibitem{canal2022survey}
Canal, F.Z., M{\"u}ller, T.R., Matias, J.C., Scotton, G.G., de~Sa~Junior, A.R., Pozzebon, E., Sobieranski, A.C.: A survey on facial emotion recognition techniques: A state-of-the-art literature review. Information Sciences  \textbf{582},  593--617 (2022)

\bibitem{dalal2005histograms}
Dalal, N., Triggs, B.: Histograms of oriented gradients for human detection. In: 2005 IEEE computer society conference on computer vision and pattern recognition (CVPR'05). vol.~1, pp. 886--893. Ieee (2005)

\bibitem{dong2024bi}
Dong, R., Lam, K.M.: Bi-center loss for compound facial expression recognition. IEEE Signal Processing Letters  (2024)

\bibitem{guo2018dominant}
Guo, J., Lei, Z., Wan, J., Avots, E., Hajarolasvadi, N., Knyazev, B., Kuharenko, A., Junior, J.C.S.J., Bar{\'o}, X., Demirel, H., et~al.: Dominant and complementary emotion recognition from still images of faces. IEEE Access  \textbf{6},  26391--26403 (2018)

\bibitem{hasani2017facial}
Hasani, B., Mahoor, M.H.: Facial expression recognition using enhanced deep 3d convolutional neural networks. In: Proceedings of the IEEE conference on computer vision and pattern recognition workshops. pp. 30--40 (2017)

\bibitem{he2022masked}
He, K., Chen, X., Xie, S., Li, Y., Doll{\'a}r, P., Girshick, R.: Masked autoencoders are scalable vision learners. In: Proceedings of the IEEE/CVF conference on computer vision and pattern recognition. pp. 16000--16009 (2022)

\bibitem{he2022compound}
He, S., Zhao, H., Yu, L., Xiang, J., Du, C., Jing, J.: Compound facial expression recognition with multi-domain fusion expression based on adversarial learning. In: 2022 IEEE International Conference on Systems, Man, and Cybernetics (SMC). pp. 688--693. IEEE (2022)

\bibitem{houssein2022human}
Houssein, E.H., Hammad, A., Ali, A.A.: Human emotion recognition from eeg-based brain--computer interface using machine learning: a comprehensive review. Neural Computing and Applications  \textbf{34}(15),  12527--12557 (2022)

\bibitem{kollias2023multi}
Kollias, D.: Multi-label compound expression recognition: C-expr database \& network. In: Proceedings of the IEEE/CVF Conference on Computer Vision and Pattern Recognition. pp. 5589--5598 (2023)

\bibitem{kollias2020analysing}
Kollias, D., Schulc, A., Hajiyev, E., Zafeiriou, S.: Analysing affective behavior in the first abaw 2020 competition. In: 2020 15th IEEE International Conference on Automatic Face and Gesture Recognition (FG 2020). pp. 637--643. IEEE (2020)

\bibitem{kollias2019face}
Kollias, D., Sharmanska, V., Zafeiriou, S.: Face behavior a la carte: Expressions, affect and action units in a single network. arXiv preprint arXiv:1910.11111  (2019)

\bibitem{kollias2021distribution}
Kollias, D., Sharmanska, V., Zafeiriou, S.: Distribution matching for heterogeneous multi-task learning: a large-scale face study. arXiv preprint arXiv:2105.03790  (2021)

\bibitem{kollias2023abaw2}
Kollias, D., Tzirakis, P., Baird, A., Cowen, A., Zafeiriou, S.: Abaw: Valence-arousal estimation, expression recognition, action unit detection \& emotional reaction intensity estimation challenges. In: Proceedings of the IEEE/CVF Conference on Computer Vision and Pattern Recognition. pp. 5888--5897 (2023)

\bibitem{kollias2023abaw}
Kollias, D., Tzirakis, P., Baird, A., Cowen, A., Zafeiriou, S.: Abaw: Valence-arousal estimation, expression recognition, action unit detection \& emotional reaction intensity estimation challenges. In: Proceedings of the IEEE/CVF Conference on Computer Vision and Pattern Recognition. pp. 5888--5897 (2023)

\bibitem{kollias20246th}
Kollias, D., Tzirakis, P., Cowen, A., Zafeiriou, S., Shao, C., Hu, G.: The 6th affective behavior analysis in-the-wild (abaw) competition. arXiv preprint arXiv:2402.19344  (2024)

\bibitem{kollias2019deep}
Kollias, D., Tzirakis, P., Nicolaou, M.A., Papaioannou, A., Zhao, G., Schuller, B., Kotsia, I., Zafeiriou, S.: Deep affect prediction in-the-wild: Aff-wild database and challenge, deep architectures, and beyond. International Journal of Computer Vision pp. 1--23 (2019)

\bibitem{kollias2019expression}
Kollias, D., Zafeiriou, S.: Expression, affect, action unit recognition: Aff-wild2, multi-task learning and arcface. arXiv preprint arXiv:1910.04855  (2019)

\bibitem{kollias2021analysing}
Kollias, D., Zafeiriou, S.: Analysing affective behavior in the second abaw2 competition. In: Proceedings of the IEEE/CVF International Conference on Computer Vision. pp. 3652--3660 (2021)

\bibitem{kollias20247th}
Kollias, D., Zafeiriou, S., Kotsia, I., Dhall, A., Ghosh, S., Shao, C., Hu, G.: 7th abaw competition: Multi-task learning and compound expression recognition. arXiv preprint arXiv:2407.03835  (2024)

\bibitem{kosti2017emotion}
Kosti, R., Alvarez, J.M., Recasens, A., Lapedriza, A.: Emotion recognition in context. In: Proceedings of the IEEE conference on computer vision and pattern recognition. pp. 1667--1675 (2017)

\bibitem{li2021facial}
Li, B., Lima, D.: Facial expression recognition via resnet-50. International Journal of Cognitive Computing in Engineering  \textbf{2},  57--64 (2021)

\bibitem{li2017reliable}
Li, S., Deng, W., Du, J.: Reliable crowdsourcing and deep locality-preserving learning for expression recognition in the wild. In: 2017 IEEE Conference on Computer Vision and Pattern Recognition (CVPR). pp. 2584--2593. IEEE (2017)

\bibitem{li2024temporallabelhierachicalnetwork}
Li, S., Lian, H., Lu, C., Zhao, Y., Qi, T., Yang, H., Zong, Y., Zheng, W.: Temporal label hierachical network for compound emotion recognition (2024), \url{https://arxiv.org/abs/2407.12973}

\bibitem{liu2023bavs}
Liu, C., Li, P., Zhang, H., Li, L., Huang, Z., Wang, D., Yu, X.: Bavs: bootstrapping audio-visual segmentation by integrating foundation knowledge. arXiv preprint arXiv:2308.10175  (2023)

\bibitem{liu2023audio}
Liu, C., Li, P.P., Qi, X., Zhang, H., Li, L., Wang, D., Yu, X.: Audio-visual segmentation by exploring cross-modal mutual semantics. In: Proceedings of the 31st ACM International Conference on Multimedia. pp. 7590--7598 (2023)

\bibitem{liu2024affectivebehaviouranalysisprogressive}
Liu, C., Zhang, W., Qiu, F., Li, L., Yu, X.: Affective behaviour analysis via progressive learning (2024), \url{https://arxiv.org/abs/2407.16945}

\bibitem{liu2024compoundexpressionrecognitionmulti}
Liu, X., Shen, K., Yao, J., Wang, B., Liu, M., An, L., Cui, Z., Feng, W., Sun, X.: Compound expression recognition via multi model ensemble for the abaw7 challenge (2024), \url{https://arxiv.org/abs/2407.12257}

\bibitem{CelebA}
Liu, Z., Luo, P., Wang, X., Tang, X.: Deep learning face attributes in the wild. In: Proceedings of the IEEE international conference on computer vision. pp. 3730--3738 (2015)

\bibitem{mollahosseini2017affectnet}
Mollahosseini, A., Hasani, B., Mahoor, M.H.: Affectnet: A database for facial expression, valence, and arousal computing in the wild. IEEE Transactions on Affective Computing  \textbf{10}(1),  18--31 (2017)

\bibitem{movellan2002tutorial}
Movellan, J.R.: Tutorial on gabor filters. Open source document  \textbf{40},  1--23 (2002)

\bibitem{nguyen2023transformer}
Nguyen, D.K., Ho, N.H., Pant, S., Yang, H.J.: A transformer-based approach to video frame-level prediction in affective behaviour analysis in-the-wild. arXiv preprint arXiv:2303.09293  (2023)

\bibitem{pietikainen2010local}
Pietik{\"a}inen, M.: Local binary patterns. Scholarpedia  \textbf{5}(3), ~9775 (2010)

\bibitem{qawaqneh2017deep}
Qawaqneh, Z., Mallouh, A.A., Barkana, B.D.: Deep convolutional neural network for age estimation based on vgg-face model. arXiv preprint arXiv:1709.01664  (2017)

\bibitem{qi2023emotiongesture}
Qi, X., Liu, C., Li, L., Hou, J., Xin, H., Yu, X.: Emotiongesture: Audio-driven diverse emotional co-speech 3d gesture generation. arXiv preprint arXiv:2305.18891  (2023)

\bibitem{qi2023diverse}
Qi, X., Liu, C., Sun, M., Li, L., Fan, C., Yu, X.: Diverse 3d hand gesture prediction from body dynamics by bilateral hand disentanglement. In: Proceedings of the IEEE/CVF Conference on Computer Vision and Pattern Recognition. pp. 4616--4626 (2023)

\bibitem{richet2024textfeaturebasedmodelscompound}
Richet, N., Belharbi, S., Aslam, H., Schadt, M.E., González-González, M., Cortal, G., Koerich, A.L., Pedersoli, M., Finkel, A., Bacon, S., Granger, E.: Text- and feature-based models for compound multimodal emotion recognition in the wild (2024), \url{https://arxiv.org/abs/2407.12927}

\bibitem{ritzhaupt2021meta}
Ritzhaupt, A.D., Huang, R., Sommer, M., Zhu, J., Stephen, A., Valle, N., Hampton, J., Li, J.: A meta-analysis on the influence of gamification in formal educational settings on affective and behavioral outcomes. Educational Technology Research and Development  \textbf{69}(5),  2493--2522 (2021)

\bibitem{IMDB-WIKI}
Rothe, R., Timofte, R., Van~Gool, L.: Deep expectation of real and apparent age from a single image without facial landmarks. International Journal of Computer Vision  \textbf{126}(2),  144--157 (2018)

\bibitem{savchenko2021facial}
Savchenko, A.V.: Facial expression and attributes recognition based on multi-task learning of lightweight neural networks. In: 2021 IEEE 19th International Symposium on Intelligent Systems and Informatics (SISY). pp. 119--124. IEEE (2021)

\bibitem{savchenko2024hsemotionteam7thabaw}
Savchenko, A.V.: Hsemotion team at the 7th abaw challenge: Multi-task learning and compound facial expression recognition (2024), \url{https://arxiv.org/abs/2407.13184}

\bibitem{savchenko2022classifying}
Savchenko, A.V., Savchenko, L.V., Makarov, I.: Classifying emotions and engagement in online learning based on a single facial expression recognition neural network. IEEE Transactions on Affective Computing  \textbf{13}(4),  2132--2143 (2022)

\bibitem{she2021dive}
She, J., Hu, Y., Shi, H., Wang, J., Shen, Q., Mei, T.: Dive into ambiguity: Latent distribution mining and pairwise uncertainty estimation for facial expression recognition. In: Proceedings of the IEEE/CVF conference on computer vision and pattern recognition. pp. 6248--6257 (2021)

\bibitem{wang2020suppressing}
Wang, K., Peng, X., Yang, J., Lu, S., Qiao, Y.: Suppressing uncertainties for large-scale facial expression recognition. In: Proceedings of the IEEE/CVF conference on computer vision and pattern recognition. pp. 6897--6906 (2020)

\bibitem{CASIA-Webface}
Yi, D., Lei, Z., Liao, S., Li, S.Z.: Learning face representation from scratch. arXiv preprint arXiv:1411.7923  (2014)

\bibitem{yin2023multi}
Yin, Y., Tran, M., Chang, D., Wang, X., Soleymani, M.: Multi-modal facial action unit detection with large pre-trained models for the 5th competition on affective behavior analysis in-the-wild. arXiv preprint arXiv:2303.10590  (2023)

\bibitem{yu2018face}
Yu, X., Fernando, B., Ghanem, B., Porikli, F., Hartley, R.: Face super-resolution guided by facial component heatmaps. In: Proceedings of the European conference on computer vision (ECCV). pp. 217--233 (2018)

\bibitem{yu2018super}
Yu, X., Fernando, B., Hartley, R., Porikli, F.: Super-resolving very low-resolution face images with supplementary attributes. In: Proceedings of the IEEE conference on computer vision and pattern recognition. pp. 908--917 (2018)

\bibitem{yu2017face}
Yu, X., Porikli, F.: Face hallucination with tiny unaligned images by transformative discriminative neural networks. In: Proceedings of the AAAI conference on artificial intelligence. vol.~31 (2017)

\bibitem{yu2017hallucinating}
Yu, X., Porikli, F.: Hallucinating very low-resolution unaligned and noisy face images by transformative discriminative autoencoders. In: Proceedings of the IEEE conference on computer vision and pattern recognition. pp. 3760--3768 (2017)

\bibitem{yu2019can}
Yu, X., Shiri, F., Ghanem, B., Porikli, F.: Can we see more? joint frontalization and hallucination of unaligned tiny faces. IEEE transactions on pattern analysis and machine intelligence  \textbf{42}(9),  2148--2164 (2019)

\bibitem{yue2019survey}
Yue, L., Chen, W., Li, X., Zuo, W., Yin, M.: A survey of sentiment analysis in social media. Knowledge and Information Systems  \textbf{60},  617--663 (2019)

\bibitem{yun2019cutmix}
Yun, S., Han, D., Oh, S.J., Chun, S., Choe, J., Yoo, Y.: Cutmix: Regularization strategy to train strong classifiers with localizable features. In: Proceedings of the IEEE/CVF international conference on computer vision. pp. 6023--6032 (2019)

\bibitem{zafeiriou2017aff}
Zafeiriou, S., Kollias, D., Nicolaou, M.A., Papaioannou, A., Zhao, G., Kotsia, I.: Aff-wild: Valence and arousal ‘in-the-wild’challenge. In: Computer Vision and Pattern Recognition Workshops (CVPRW), 2017 IEEE Conference on. pp. 1980--1987. IEEE (2017)

\bibitem{zhang2017mixup}
Zhang, H., Cisse, M., Dauphin, Y.N., Lopez-Paz, D.: mixup: Beyond empirical risk minimization. arXiv preprint arXiv:1710.09412  (2017)

\bibitem{zhang2022multi}
Zhang, T., Liu, C., Liu, X., Liu, Y., Meng, L., Sun, L., Jiang, W., Zhang, F., Zhao, J., Jin, Q.: Multi-task learning framework for emotion recognition in-the-wild. In: European Conference on Computer Vision. pp. 143--156. Springer (2022)

\bibitem{zhang2023multi}
Zhang, W., Ma, B., Qiu, F., Ding, Y.: Multi-modal facial affective analysis based on masked autoencoder. In: Proceedings of the IEEE/CVF Conference on Computer Vision and Pattern Recognition. pp. 5792--5801 (2023)

\bibitem{zhang2024effective}
Zhang, W., Qiu, F., Liu, C., Li, L., Du, H., Guo, T., Yu, X.: An effective ensemble learning framework for affective behaviour analysis. In: Proceedings of the IEEE/CVF Conference on Computer Vision and Pattern Recognition. pp. 4761--4772 (2024)

\bibitem{zhang2024affectivebehaviouranalysisintegrating}
Zhang, W., Qiu, F., Liu, C., Li, L., Du, H., Guo, T., Yu, X.: Affective behaviour analysis via integrating multi-modal knowledge (2024), \url{https://arxiv.org/abs/2403.10825}

\bibitem{zhang2022learn}
Zhang, Y., Wang, C., Ling, X., Deng, W.: Learn from all: Erasing attention consistency for noisy label facial expression recognition. In: European Conference on Computer Vision. pp. 418--434. Springer (2022)

\bibitem{zhu2021webface260m}
Zhu, Z., Huang, G., Deng, J., Ye, Y., Huang, J., Chen, X., Zhu, J., Yang, T., Lu, J., Du, D., et~al.: Webface260m: A benchmark unveiling the power of million-scale deep face recognition. In: Proceedings of the IEEE/CVF Conference on Computer Vision and Pattern Recognition. pp. 10492--10502 (2021)

\end{thebibliography}
\end{document}